\documentclass[twocolumn, switch]{article} 

\usepackage{preprint}

\usepackage{amsmath, amsthm, amssymb, amsfonts}

\usepackage[numbers,square]{natbib}
\usepackage[utf8]{inputenc}	
\usepackage[T1]{fontenc}	
\usepackage{xcolor}		
\usepackage[colorlinks = true,
            linkcolor = purple,
            urlcolor  = blue,
            citecolor = cyan,
            anchorcolor = black]{hyperref}	
\usepackage{booktabs} 		
\usepackage{nicefrac}		
\usepackage{microtype}		
\usepackage{lineno}		
\usepackage{float}			
\usepackage{csvsimple}
\usepackage{siunitx}

\usepackage{float}
\usepackage{listings}

\newfloat{lstfloat}{htbp}{lop}
\floatname{lstfloat}{Listing}

\usepackage{newfloat}
\DeclareFloatingEnvironment[name={Supplementary Figure}]{suppfigure}
\usepackage{sidecap}
\sidecaptionvpos{figure}{c}

\usepackage{titlesec}
\titlespacing\section{0pt}{12pt plus 3pt minus 3pt}{1pt plus 1pt minus 1pt}
\titlespacing\subsection{0pt}{10pt plus 3pt minus 3pt}{1pt plus 1pt minus 1pt}
\titlespacing\subsubsection{0pt}{8pt plus 3pt minus 3pt}{1pt plus 1pt minus 1pt}

\usepackage{tikz,xcolor,hyperref}
\usepackage{numprint}

\definecolor{lime}{HTML}{A6CE39}
\DeclareRobustCommand{\orcidicon}{
	\begin{tikzpicture}
	\draw[lime, fill=lime] (0,0) 
	circle [radius=0.16] 
	node[white] {{\fontfamily{qag}\selectfont \tiny ID}};
	\draw[white, fill=white] (-0.0625,0.095) 
	circle [radius=0.007];
	\end{tikzpicture}
	\hspace{-2mm}
}
\foreach \x in {A, ..., Z}{\expandafter\xdef\csname orcid\x\endcsname{\noexpand\href{https://orcid.org/\csname orcidauthor\x\endcsname}
			{\noexpand\orcidicon}}
}
\title{Intensive Care as One Big Sequence Modeling Problem}

\usepackage{authblk}

\author[1\thanks{\tt{v.liventsev@tue.nl}}]{Vadim Liventsev}
\author[2\thanks{\tt{tobias.fritz@unibw.de}}]{Tobias Fritz}

\affil[1]{Department of Mathematics and Computer Science, Eindhoven University of Technology}
\affil[2]{Research Institute CODE, Bundeswehr University Munich}

\begin{document}

\twocolumn[ 
  \begin{@twocolumnfalse} 
  
\maketitle

\begin{abstract}
Reinforcement Learning in Healthcare is typically concerned with narrow self-contained tasks such as sepsis prediction or anesthesia control.
However, previous research has demonstrated the potential of generalist models (the prime example being Large Language Models) to outperform task-specific approaches due to their capability for implicit transfer learning.
To enable training of foundation models for Healthcare as well as leverage the capabilities of state of the art Transformer architectures, we propose the paradigm of Healthcare as Sequence Modeling, in which interaction between the patient and the healthcare provider is represented as an event stream and tasks like diagnosis and treatment selection are modeled as prediction of future events in the stream. 
To explore this paradigm experimentally we develop MIMIC-IV-Ext-SEQ, a sequence modeling benchmark derived by translating heterogenous clinical records from MIMIC-IV dataset into a uniform event stream format, train a baseline model and explore its capabilities.
\end{abstract}
\vspace{0.35cm}

  \end{@twocolumnfalse} 
] 



\footnotetext{The dataset can be downloaded at \url{https://physionet.org/content/MIMIC-IV-Ext-SEQ/}.
More instuctions as well as standard evaluation scripts can be found at \url{https://github.com/vadim0x60/mimicseq}. 
Implementation of the baseline model can be found at \url{https://github.com/Teywazz/irregular_time_series}}

\section{Introduction}


\subsection{Intensive Care databases: a nano-review}
\label{sec:datasets}

Intensive Care is the medical speciality that supports patients whose lives are in immediate danger.
As such, it requires robust real time monitoring of the patient's vital signs due to quickly identify potential deterioration \citep{Bailey2013trial, Blount2010Real, Bockholt2022Real, Dimitrios1999Distributed, Fried2000Some, Mao2012integrated, Prgomet2016Vital, Vincent2018Improving}.
Monitoring hardware creates a datastream of variables like heart rate and blood oxygenation and, as a result, Intensive Care stands to benefit more than other fields of Healthcare from integration of data-driven models \citep{nunezreizBigDataAnalysis2019}.

MIMIC IV \citep{johnsonMIMICIVFreelyAccessible2023}, AmsterdamUMCdb \citep{amsterdamumcdb-a}, HiRID \citep{yecheHiRIDICUBenchmarkComprehensiveMachine} and the eICU Collaborative Research Database \cite{pollard2018a} are databases of health records obtained from Intensive Care Units.
Unlike many machine learning datasets, they avoid setting a standard for the parameters of the task such as which variables of a sample are to be used as features and which are to be predicted by the model, which samples are to be used for training and which are the holdout set - even what is a sample (a patient? an ICU admission? a drug? a diagnosis?). This makes them flexible and suitable for a wide array of tasks, but presents a challenge when seeking to compare different studies \cite{mcdermottReproducibilityMachineLearning2021}.

Derivative benchmarks, such as the MIMIC Benchmark \cite{harutyunyanMultitaskLearningBenchmarking2019} and HiRID-ICU-Benchmark \citep{yecheHiRIDICUBenchmarkComprehensiveMachine} seek to address the reproducibility issue: they put forth multiple datasets (one for each task) where all the information is derived from MIMIC and HiRID-ICU respectively, but is arranged specifically for the learning task at hand. They aim to become standard benchmarks for the tasks of mortality and length of stay prediction, patient phenotyping, prediction of circulatory, respiratory or kidney failure.

\subsection{Towards foundation models for Healthcare}

Generalist models have demonstrated a superior performance to task specific models in many areas of machine learning \cite{reedGeneralistAgent2022} due to their ability to exploit implicit shared subtasks. 
This finding has precipitated the birth of a new paradigm known as foundation models \cite{zhouComprehensiveSurveyPretrained2023} - models trained on an all-encompassing dataset (such as the dataset that attempts to approximate all written text \cite{chelbaOneBillionWord2013}) and designed to be adapted to a broad array of specific downstream tasks.

The learning tasks typically studied in Healthcare are often interrelated. 
To use an example from section \ref{sec:datasets}, sepsis presents a high risk of death \cite{schlichtingRecognizingManagingSevere2007} and predicting sepsis is evidently useful for predicting mortality.
In light of existing research on generalist models, it is likely that considering them separately is counterproductive.
This, together with the vital societal importance of access to healthcare amidst understaffing \citep{ashleyy.metcalfHospitalUnitUnderstaffing2016, hudsonUnderstaffing2015, mercerMessageEditorinChief2008, munnUnderstaffingWardsCompromising2017, r.stanleyUnderstaffedOverwhelmed2010, SurveyShowsHidden1993, thelancetHealthcareSystemStaffing2018, UnderstaffingSignificantIssue2012} and population ageing \citep{2012health, Aslam2021Ageing, L1991aging, Lloyd2012Population, Mahishale2015Ageing, Mann2004aging, Sammy2019global, Suzman2015Health}, makes a foundation model for Healthcare a particularly important research goal.

\subsection{The case for Healthcare as Sequence Modeling}
\label{sec:sequencemodel}

With a few notable exceptions foundation models tend to be sequence models: estimating the probability a sequence of fixed-size elements

\begin{equation}
    \hat{p}(t_1, \dots, t_n)
\end{equation}

typically modeled as conditional probability of one (usually last) element given others

\begin{equation}
    \hat{p}(t_n | t_1, \dots, t_n)
\end{equation}

since most tasks in machine learning can be represented as sequence modeling. 
There is evidence that this generality is a fundamental property of sequence models: see proofs that Recurrent Neural Networks \cite{siegelmannComputationTuringLimit1995} and Attention \cite{perezAttentionTuringComplete} are Turing-complete.

One of the tasks that has recently been reimagined as a Sequence Modeling task is Reinforcement Learning and Healthcare has a Reinforcement Learning interpretation: it's an interaction between the doctor (the agent) and the patient's body (the environment) that has all the trappings of a POMDP: the treatment interventions are actions $a_n$ and the observable vital signs are observations $o_n$.
Consider a \emph{trajectory} in a Partially Observable Markov Decision Process, i.e. a history of actions $a$, observations $o$ and rewards $r$:

\begin{equation}
    \tau = (o_1, a_1, r_1, o_2, a_2, r_2, o_3, a_3, r_3, \dots)
\end{equation}

A model $\hat{p}$ that can predict the next element in a trajectory can be used as a \emph{dynamics model} to predict the next observation from the patient:

\begin{equation}
    o_{n+1} \sim \hat{p}(o_{n+1} | \dots, o_n, a_n, r_n)
\end{equation}

as an \emph{imitative policy} to predict which intervention the doctor will choose next

\begin{equation}
    a_{n+1} \sim \hat{p}(a_{n+1} | \dots a_n, r_n, o_{n+1})
\end{equation}

or, after an additional optimization step, as an \emph{optimal policy} to predict which intervention the doctor \emph{should} choose next.
For a planning horizon of 1 (greedy policy) it is defined as:

\begin{equation}
    a_{n+1} = \arg \max_a \mathbb{E} [r_{n+1} \sim \hat{p}(r_{n+1} | \dots a_n, r_n, o_{n+1}, a)]
\end{equation}

Thus while the paradigm of Healthcare as a Sequence Modeling task is novel, there is a robust body of research reducing Healthcare to Reinforcement Learning \cite{yuReinforcementLearningHealthcare2021} and Reinforcement Learning to Sequence Modeling \cite{chenDecisionTransformerReinforcement2021, jannerOfflineReinforcementLearning2021, schmidhuberReinforcementLearningUpside2020}. 

\subsection{Contributions}

In this paper we introduce
\begin{itemize}
    \item the paradigm of Healthcare as Sequence Modeling
    \item MIMIC-Ext-SEQ: a benchmark dataset for Sequence Models of Intensive Care derived from MIMIC-IV
    \item Evaluation guidelines for a holistic comparison of intensive care forecasting models
    \item An MLP-based baseline model
\end{itemize}

\section{Related Work}

\citet{weiMIMICELMIMICIVEvent2022} develop MIMICEL, a linearized event sequence representation of MIMIC-IV-ED \cite{johnsonMIMICIVED2021} emergency department dataset for downstream application in machine learning and process mining. 
However, MIMIC-IV-ED is a smaller dataset than MIMIC-IV, and \citet{weiMIMICELMIMICIVEvent2022} stop short of releasing a full benchmark suite with an evaluation procedure and baselines.
Incorporating MIMICEL into MIMIC-Ext-SEQ would be a fruitful direction of future research.
Similarly to our approach,
\cite{kuznetsovaImportanceStepwiseEmbeddings2023} and \cite{tipirneniSelfSupervisedTransformerSparse2022} represent intensive care data as an irregular event stream timeseries, however they still rely on manual selection of important variables and prediction targets.

\section{MIMIC-Ext-SEQ}
\label{sec:dataset}

MIMIC-Ext-SEQ is a dataset for training foundational models for Intensive Care in the Sequence Modeling paradigm derived from MIMIC-IV.
It sets out a single machine learning task, with fully standardized train and test sets for reproducible comparisons between methods, while, at the same time, the task is so general that if a model accomplishes it successfully, this model can be used without fine-tuning for many narrow tasks, including mortality and length of stay prediction.
We also provide a suite of evaluation metrics and 2 baseline models.


\subsection{Building the timeseries dataset}

For every patient admission recorded in MIMIC-IV we collect all related information from various heterogeneous subdatasets, namely,

\begin{itemize}
    \item ICU input events (\texttt{inputevents})
    \item ICU procedures (\texttt{procedureevents})
    \item Hospital prescriptions (\texttt{prescriptions})
    \item Patient admissions and demographics (\texttt{admissions} and \texttt{patients})
    \item ICU charted events (\texttt{chartevents})
    \item Hospital lab events (\texttt{labevents})
    \item Microbiology tests (\texttt{microbiologyevents})
    \item Procedures coded in ICD format (\texttt{procedures\_icd})
    \item Healthcare Common Procedure Coding System (HCPCS) events (\texttt{hcpcsevents})
    \item Electronic medication administration records (eMAR) (\texttt{emar})
\end{itemize}

Every patient history is then represented as a sequence of events where each event has:
\begin{enumerate}
    \item a type (each type has an associated text label, i.e, "Penatal given")
    \item time when it happened
    \item (optionally) intensity
\end{enumerate}

Intensities may represent dosages of drugs or other quantitative measures (charted heart rate, blood pressure, patient age).

Events that have a duration, such as medication administrations, are recorded as 2 events: "start X" and "end X". 
Demographic information, including ethnicity, gender, and age, is recorded as a dummy event occurring the time of admission.

\subsection{Resulting dataset}

\begin{table*}
    \centering
    \begin{tabular}{rlrrr}
\toprule
 event\_id &                                          label &  intensity\_mean &  intensity\_std &  frequency \\
\midrule
    22317 &                                Safety Measures &             NaN &            NaN &    9592332 \\
     8534 &                                     Heart Rate &           87.78 &        3837.73 &    6798187 \\
    14229 &                               Respiratory Rate &           20.45 &         915.30 &    6728530 \\
    10787 &                    O2 saturation pulseoxymetry &          104.57 &        7961.00 &    6656949 \\
    25364 &                                   Heart Rhythm &             NaN &            NaN &    6220746 \\
    31399 &                                  Ectopy Type 1 &             NaN &            NaN &    5553201 \\
    14230 &           Non Invasive Blood Pressure systolic &          120.59 &         858.58 &    4279569 \\
      564 &          Non Invasive Blood Pressure diastolic &           66.35 &         306.11 &    4278628 \\
    11908 &               Non Invasive Blood Pressure mean &           84.25 &        5260.15 &    4276928 \\
    31400 &                      Less Restrictive Measures &             NaN &            NaN &    3505880 \\
     3981 &                                          Foley &          119.56 &         129.90 &    3137879 \\
    43540 &                                    Head of Bed &             NaN &            NaN &    2569735 \\
    52374 &                                           Turn &             NaN &            NaN &    2536571 \\
     9644 &                                     Hemoglobin &           34.68 &        4953.36 &    2449532 \\
    40529 &                             Activity Tolerance &             NaN &            NaN &    2437173 \\
    86048 &                                    Orientation &            1.00 &           0.00 &    2403976 \\
    49436 &                                       Position &             NaN &            NaN &    2395796 \\
     4599 &                   Arterial Blood Pressure mean &           80.10 &         631.99 &    2387853 \\
     5748 &               Arterial Blood Pressure systolic &          119.34 &          86.83 &    2379566 \\
     4598 &              Arterial Blood Pressure diastolic &           60.04 &         259.39 &    2379199 \\
    13671 &                                 Platelet Count &          247.16 &        5160.22 &    2336562 \\
     1147 &                                        Glucose &          133.10 &         876.23 &    2325179 \\
     7921 &                                      Magnesium &           11.93 &        3142.74 &    2227639 \\
    31353 &                                Pain Management &             NaN &            NaN &    2127823 \\
    10227 &                                      Potassium &            4.14 &           0.60 &    2078644 \\
     6880 &                                         Sodium &          138.38 &           4.78 &    2058719 \\
    13749 &                                       Chloride &          102.20 &           5.91 &    2042431 \\
     3467 &                                     Hematocrit &           31.12 &           6.22 &    2041709 \\
    55316 &            Sodium Chloride 0.9\%  Flush Flushed &             NaN &            NaN &    2002785 \\
     1197 &                                     Creatinine &            1.43 &           1.53 &    1995732 \\
    11416 &                                    Bicarbonate &           25.38 &           4.51 &    1987133 \\
    40492 & Altered Respiratory Status NCP - Interventions &             NaN &            NaN &    1986935 \\
     3442 &                                      Anion Gap &           13.89 &           3.55 &    1981798 \\
     5733 &                                  Urea Nitrogen &           25.93 &          21.62 &    1981280 \\
    49437 &                               Temperature Site &             NaN &            NaN &    1937625 \\
     1149 &                              White Blood Cells &            9.64 &          15.42 &    1890342 \\
     1187 &                                           MCHC &           32.77 &           1.66 &    1886210 \\
     1195 &                                Red Blood Cells &            3.45 &           0.73 &    1886104 \\
    14805 &                                            MCH &           29.86 &           2.82 &    1886103 \\
    14806 &                                            MCV &           91.17 &           7.43 &    1886103 \\
    11415 &                                            RDW &           15.59 &           2.59 &    1886033 \\
    24949 &                                Therapeutic Bed &             NaN &            NaN &    1815296 \\
    11716 &                              GCS - Eye Opening &            3.30 &           1.04 &    1711688 \\
     9422 &                          GCS - Verbal Response &            3.19 &           1.87 &    1708459 \\
     3518 &                           GCS - Motor Response &            5.30 &           1.42 &    1704243 \\
    13701 &                                      Phosphate &            3.57 &           1.17 &    1685984 \\
     2361 &                                 Calcium, Total &            8.62 &           0.77 &    1685869 \\
    49039 &                                Education Topic &             NaN &            NaN &    1634871 \\
    46053 &                                   Pain Present &             NaN &            NaN &    1634416 \\
    37045 &                                     Assistance &             NaN &            NaN &    1633605 \\
\bottomrule
\end{tabular}
    \label{tab:top-events}
    \caption{50 most frequent event types}
\end{table*}

\begin{table*}[]
    \centering
    \begin{tabular}{crl}
    \toprule
    eventtime &	label &	intensity \\
    \midrule
    2185-08-13T16:57:00 &	WHITE	 & \\
    2185-08-13T16:57:00 &	AGE &	18.0 \\
    2185-08-13T16:57:00 &	URGENT ADMISSION &	 \\
    2185-08-13T16:57:00 &	FEMALE &	 \\
    2185-08-13T21:00:00 &	Start Prenatal Vitamins Tablet prescription, PO &	1.0 \\
    2185-08-13T22:00:00 &	Start LORazepam 1mg Tablet prescription, PO/NG &	 \\
    2185-08-13T22:00:00 &	Start HydrOXYzine 25 mg Tab prescription, PO/NG &	 \\
    2185-08-14T08:00:00 &	Start Complera 200 mg-25 mg-300 mg tablet prescription, ORAL &	1.0 \\
    2185-08-14T17:00:00 &	Start Acetaminophen 325mg Tablet prescription, PO/NG &	 \\
    2185-08-19T11:30:00 &	DISCHARGE TO HOME &	 \\
    2185-08-19T18:00:00 &	Stop LORazepam 1mg Tablet prescription, PO/NG &	 \\
    2185-08-19T18:00:00 &	Stop Prenatal Vitamins Tablet prescription, PO &	1.0 \\
    2185-08-19T18:00:00 &	Stop HydrOXYzine 25 mg Tab prescription, PO/NG &	 \\
    2185-08-19T18:00:00 &	Stop Complera 200 mg-25 mg-300 mg tablet prescription, ORAL &	1.0 \\
    2185-08-19T18:00:00 &	Stop Acetaminophen 325mg Tablet prescription, PO/NG &	 \\
    \bottomrule
    \end{tabular}
    \label{tab:sample-patient}
    \caption{An (unusually short) hospital admission from MIMIC-Ext-SEQ}
\end{table*}

MIMIC-Ext-SEQ contains 481374190 clinical events in 522740 train and 10000 test episodes (hospital admissions).

\begin{table}[H]
    \centering
    \begin{tabular}{|c|c|c|c|}
        \toprule
         & min & avg & max \\
         \midrule
         Events per episode & 4 & 919 & 564721 \\
         Episode duration & 0 & 6 days & 68 years \\
         \bottomrule
    \end{tabular}
    \caption{Dataset statistics}
    \label{tab:stats}
\end{table}

It is publicly available, subject to (no cost, open to everyone) MIMIC-IV data use agreement.
See the dataset repository.

\subsection{Clustering}

87899 event types is a very fine-grained view of the intensive care scenario that differentiates, for example, between different versions of the same drug (say, pills and tablets).
This is done intentionally to pave the way for sophisticated models, however, we recognize that a simplified version of the task can be helpful.
To that end, we enrich the dataset with 4 \emph{clusterings}: c10, c100, c1000 and c10000.
They are achieved by embedding each event type label with OpenAI's \emph{ada2} embedding model \cite{neelakantanTextCodeEmbeddings2022a} and running a k-means clustering algorithm in the induced latent space.
As a result, one can train a model for simplified \emph{clustered} versions the task making predictions in terms of broad event categories, not individual event types.

\subsection{Train test split}

For the purposes of model evaluation, benchmarking and leaderboards 10000 episodes (admissions) were selected randomly, stratified by admission type (walk in, physician referral, transfer from hospital, etc.), discharge type (to home, to other facility, death), ethnicity, age and gender to ensure a representative sample.
The rest of the episodes are included in the training set.

We sort the 2 sets by episode length to simplify batching.

\subsection{Evaluation guidelines}

We suggest evaluating forecasting models on 2 test tasks using the holdout patient histories:
\begin{description}
    \item[second day prediction] for every episode in the holdout set, use all events within 24 hours of the very first event as model input. Use the next 24 hours as expected output.
    \item[last day prediction] for every episode in the holdout set, use all events within 24 hours of the very last event as expected output. Use the rest of the events as model input. 
\end{description}

Each of the two can in turn be decomposed into:
\begin{description}
    \item[event prediction task] which events will happen and which will not?
    \item[intensity prediction task] if the event happens, what will be its intensity?
\end{description}

The former is a \emph{binary classification task} with metrics like \emph{accuracy}, \emph{precision} and \emph{recall}.
Note that it's a highly imbalanced binary classification task and, as such, relying on accuracy is not recommended - \emph{f1 score}, \emph{kappa} or \emph{dice score} shall be used instead.
The latter is a \emph{regression task} and the recommended metric is $R^2$ coefficient.

\emph{Event prediction} can be done in the space of concrete events or in the space of clusters c10, c100, c1000, c10000.
So, in total, a model evaluation includes 10 binary classification tasks (second day prediction and last day prediction for each event granularity) and 2 regression tasks.

\subsection{Relationship to existing benchmarks}

A predictive model trained on \emph{MIMIC-Ext-SEQ} can perform the standard tasks used in current benchmarks \cite{harutyunyanMultitaskLearningBenchmarking2019}, such as
\begin{itemize}
    \item \emph{length of stay prediction} by estimating the probability of \texttt{DISCHARGE TO HOME} \\ \texttt{DISCHARGE TO REHAB} \\ \texttt{DISCHARGE TO HEALTHCARE FACILITY} \\ \texttt{DISCHARGE TO PSYCH FACILITY}\\\texttt{DISCHARGE TO OTHER FACILITY}\\\texttt{DISCHARGE TO REHAB}\\\texttt{DISCHARGE TO ASSISTED LIVING}\\\texttt{DISCHARGE TO HOSPICE}\\\texttt{DISCHARGE TO ACUTE HOSPITAL}\\\texttt{DISCHARGE AGAINST ADVICE} \\ \texttt{DISCHARGE TO HOME HEALTH CARE} \\ \texttt{DISCHARGE TO CHRONIC/LONG TERM ACUTE CARE} \\ \texttt{DISCHARGE TO SKILLED NURSING FACILITY} \\ \texttt{DISCHARGE TO DIED} \\ events over different timescales
    \item \emph{mortality prediction} by estimating the probability of \texttt{DISCHARGE TO DIED} relative to other types of discharge
    \item \emph{decompensation prediction} by estimate the probability of \texttt{DISCHARGE TO DIED} within 24 hours and/or events known to signify an acute increase in letality
    \item \emph{phenotyping} by estimating the probability of various \texttt{DISEASE X DIAGNOSED} events
\end{itemize}

as well as many others.

\section{Baseline}
\label{sec:baseline}

Our baseline model consists of a two-layer multilayer perceptron (MLP) with 1000 hidden layer size RELU \cite{agarapDeepLearningUsing2018} activation function and batch normalization after each layer. As input, all events from the first day are used and one-hot encoded in a 87899-dimensional vector. As prediction target, all events from the second day are used and encoded via their clustering mapping, e.g. c10, c100, c1000, c10000, into a vector of the corresponding dimension. As objective function we used binary cross entropy. The last layer contains a sigmoid function which transforms the output to probabilities for each vector element. A threshold is set at 0.5 to decide if an event occurs or not. All models were trained with batch size 512 for 3 epochs.

We test our baseline on the \emph{second day event prediction task} and summarize the results for different clusterings in table \ref{tab:mytable2}. It can be seen that the more classes are in the clustering, the harder the prediction task becomes. As noted above, accuracy is a deceptive metric in this scenario.

\begin{table}[H]
  \centering
    \begin{tabular}{lcccc} \toprule
        {clustering} & {recall} & {accuracy} & {precision} & {F1}  \\ \midrule
        {c10}  & 0.903 & 0.840 & 0.790 & 0.827 \\
        {c100}  & 0.568 & 0.885 & 0.713 & 0.632 \\
        {c1000}  & 0.500  & 0.976 & 0.710  & 0.586 \\
        {c10000}  & 0.509  & 0.996 &  0.703  & 0.589 \\ \midrule
    \end{tabular}
  \caption{Evaluation results of 2 x 1000 MLP for first day - second day  prediction, entire dataset}
  \label{tab:mytable2}
\end{table}

For many patients it is the case that they are in the hospital for only one day. For these patients, the model should predict no event on the second day. However, evaluation of the models showed that this is almost never the case. However, one can argue that it is more important to get problematic patients correct than the ones who leave the hospital after one day. Therefore, we evaluated the model also only on patients which are in the hospital for at least 2 days. Since the data is roughly ordered according to length of stay, we achieved this by skipping the first 100k samples in the training and the first 1k samples in the test data. The models shows improved performance here, as can be seen in Table \ref{tab:mytable1}.

\begin{table}[H]
  \centering
    \begin{tabular}{lcccc} \toprule
        {clustering} & {recall} & {accuracy} & {precision} & {F1}  \\ \midrule
        {c10}  & 0.942  & 0.854 & 0.850  &  0.890 \\ 
        {c100}  & 0.633  & 0.878 & 0.765  &  0.692 \\ 
        {c1000}  & 0.548  & 0.974 & 0.770  &  0.640 \\ 
        {c10000}  &  0.539  & 0.995 & 0.771  &  0.634 \\ \midrule
        
    \end{tabular}
  \caption{Evaluation results of 2 x 1000 MLP for first day - second day  prediction, skipping first 100k train / 1k test samples}  \label{tab:mytable1}
\end{table}


The same setup but with 3 hidden layers and 5000 units each shows improved performance as can be seen in table \ref{tab:mytable3}. Bigger models were tested as well, but showed no further improvement.

\begin{table}[H]
  \centering
    \begin{tabular}{lcccc} \toprule
        {configuration} & {recall} & {accuracy} & {precision} & {F1}  \\ \midrule
        {1}  & 0.505  & 0.996 &  0.727  & 0.595 \\
        {2} & 0.544  & 0.9959 &  0.784  & 0.642 \\  \midrule
    \end{tabular}
  \caption{All evaluations with c10000; 1: Evaluation results of 3 x 5000 MLP for first day - second day  prediction, entire dataset; 2: Evaluation results of 3 x 5000 MLP for first day - second day prediction, skip first 100k train / 1k test samples}
  \label{tab:mytable3}
\end{table}

Replacing ones in the one-hot encoding with the corresponding events' intensities impaired the models' performance, likely because it introduces a false equivalency between a zero-intensity event and lack of an event, i.e. "average blood pressure" is different from "no blood pressure measurement".


\section{Conclusion}

Narrow tasks in machine learning for intensive care have been a result of technical limitations that have become less relevant with recent advances in the field. 
We propose a more general approach, publish a dataset to support it and demonstrate its viability with a simple baseline model.
The long term ambition of this work is to become the basis for training foundation models of intensive care using modern neural network architectures.
Of particular interest are Transformers \cite{vaswaniAttentionAllYou2023}, Neural Controlled Differential Equations \cite{kidgerNeuralControlledDifferential2020} and Structured State Space Models \cite{guEfficientlyModelingLong2022}.

\section*{Acknowledgments}

This project has received funding from European Union’s Horizon 2020 research and innovation programme under grant agreement n° 812882. 



\normalsize
\bibliography{references}

\clearpage

\end{document}